\documentclass[a4paper]{article}

\usepackage{INTERSPEECH2016}

\usepackage{graphicx}
\usepackage{amssymb,amsmath,bm}
\usepackage{textcomp}
\usepackage{color}
\usepackage{booktabs}
\usepackage{tabu}

\DeclareMathOperator*{\inc}{inconsist}
\DeclareMathOperator*{\con}{const}
\DeclareMathOperator*{\req}{req}

\ninept

\title{A Sequence-to-Sequence Model for User Simulation in Spoken Dialogue Systems}

\makeatletter
\def\name#1{\gdef\@name{#1\\}}
\makeatother
\name{{\em Layla El Asri, Jing He, Kaheer Suleman}}

\address{Maluuba Research\\
				Montreal, Canada\\
  {\small \tt first.last@maluuba.com}
}

\begin{document}

  \maketitle
  \begin{abstract}
User simulation is essential for generating enough data to train a statistical spoken dialogue system. Previous models for user simulation suffer from several drawbacks, such as the inability to take dialogue history into account, the need of rigid structure to ensure coherent user behaviour, heavy dependence on a specific domain, the inability to output several user intentions during one dialogue turn, or the requirement of a summarized action space for tractability. This paper introduces a data-driven user simulator based on an encoder-decoder recurrent neural network. The model takes as input a sequence of dialogue contexts and outputs a sequence of dialogue acts corresponding to user intentions. The dialogue contexts include information about the machine acts and the status of the user goal. We show on the Dialogue State Tracking Challenge 2 (DSTC2) dataset that the sequence-to-sequence model outperforms an agenda-based simulator and an n-gram simulator, according to F-score. Furthermore, we show how this model can be used on the original action space and thereby models user behaviour with finer granularity.

  \end{abstract}

 \section{Introduction}
Statistical Spoken Dialogue Systems (SDS) \cite{Levin:97,Gasic:11,Daubigney:12b} typically require several thousands of dialogues to learn a good dialogue strategy \cite{Pietquin:11, Gasic:12}. It is costly to collect this quantity of dialogues; therefore, research has turned to user simulation \cite{Eckert:97,Georgila:06,Schatzmann:06,Chandramohan:11}. A user simulator is expected to have the following properties: to be statistically consistent with real users, to generate coherent sequences of actions, and to generalize to new contexts \cite{Pietquin:13}. User simulation can be either at the intention level, \textit{i.e.}, generating dialogue acts \cite{Schatzmann07b,Chandramohan:11}, or at the utterance level \cite{Jung:09}. In this work, we focus on the intention level.

Many models have been designed in order to meet the requirements cited above \cite{Eckert:97,Scheffler02,Georgila05,Cuayahuitl05,Schatzmann07b,Chandramohan:11}. These models typically suffer from important drawbacks, which include the inability to take dialogue history into account \cite{Eckert:97}, the need of rigid structure to ensure coherent user behaviour \cite{Schatzmann07a}, heavy dependence on a specific domain \cite{Scheffler02}, the inability to output several user intentions during one dialogue turn \cite{Jung:09}, or the requirement of a summarized action space for tractability \cite{Schatzmann07b}.

In this paper, we introduce a sequence-to-sequence model for user simulation. The simulator is modelled with an encoder Recurrent Neural Network (RNN) and a decoder RNN \cite{Sutskever:14}. The encoder takes as input the entire history of the dialogue, encoded as a sequence of dialogue contexts. It outputs an internal representation of this sequence. This representation is passed as input to the decoder. The decoder generates a sequence of dialogue acts corresponding to user intentions.

We train this model on the Dialogue State Tracking Challenge 2 (DSTC2) \cite{Henderson14a} dataset. This corpus consists of dialogues between real users and an SDS in the domain of restaurant-seeking. We compare the sequence-to-sequence model to an agenda-based simulator \cite{Schatzmann07a,Schatzmann07b}, an n-gram simulator, and a sequence-to-one RNN which takes the same input as the sequence-to-sequence model but chooses an output from among a list of predefined sequences of acts. We show that the RNN-based models outperform the other two simulators on the F-score measure. We also show on the DSTC3 dataset \cite{Henderson:14c} that the RNN-based models generalize best to new domains.

In the next section, we discuss previous models for user simulation. Then, in Section \ref{seqtoseq}, we describe the sequence-to-sequence model. Section \ref{expes} presents the DSTC2 and DSTC3 corpora, the models and the metrics used for comparison, and the results of our experiments.

\begin{table*}
\begin{center}
\begin{tabu}to\linewidth{ccccc}
    \toprule
Machine output / User answer & Machine acts & Inconsistency vector & Constraints status & Requests status \\ \midrule
Welcome! How may I help you? & 0000000010 & 000000 & {\color{blue}0}{\color{red}0}1 & 1011{\color{magenta}0}111 \\
& greet &&& \\\cmidrule(rl){1-5}
Is there a {\color{blue}cheap} restaurant {\color{red}downtown}? &&&&\\\cmidrule(rl){1-5}
A {\color{blue}cheese} restaurant. & 0000010001 & 0000{\color{blue}1}0 & {\color{blue}0}{\color{red}1}1 & 1011{\color{magenta}0}111 \\
What is your budget? & implicit-confirm, request &&& \\\cmidrule(rl){1-5}
No, I said a cheap restaurant. &&&&\\\cmidrule(lr){1-5}
Panda express is a cheap &  0100000100 & 000000 & {\color{blue}1}{\color{red}1}1 & 1011{\color{magenta}0}111 \\
restaurant downtown. & offer, inform &&& \\\cmidrule(lr){1-5}
What is the {\color{magenta}address} of this place? &&&&\\\cmidrule(rl){1-5}
Panda express is located &  0100000100 & 000000 & {\color{blue}1}{\color{red}1}1 & 1011{\color{magenta}1}111 \\
at 108 Queen street. & offer, inform &&& \\ \bottomrule
\end{tabu}
\caption{Examples of contexts in a dialogue with a restaurant-seeking system. The user goal has two constraints (cheap and downtown) and one request (address).\vspace*{-2mm}}
\end{center}
\label{tab:ex_contexts}
\end{table*}

 \section{Background}
User simulation, at the intention level, consists of predicting the next user dialogue act depending on the dialogue history and the user goal. The first user simulator was proposed by Eckert et al. \cite{Eckert:97} who used a simple bi-gram model $P(a_u | a_m)$ to predict the next user act $a_u$ given the last system act $a_m$. This model does not produce coherent behaviours from the user because the user only reacts to the latest system action. This issue can be overcome by restricting the types of actions that the user can draw from according to dialogue history, which requires more engineering effort. Scheffler and Young \cite{Scheffler02} proposed a graph-based model. Therein, all possible paths for user behaviour are mapped into a network. The main difficulty of this approach is that it requires extensive domain knowledge and engineering. Pietquin and Dutoit \cite{Pietquin06} suggested a Bayesian model for user behaviour. They added an explicit representation of the user goal and memory to the probabilistic bi-gram model. The user's action was then conditioned on her goal and memory. Georgila et al. \cite{Georgila05} proposed a richer model of the user with the information state approach \cite{Larsson:00}. The information state carries information on the current state, the dialogue history and ongoing actions. The authors investigated learning user behaviour by using a 4-gram representation and a linear combination to map each state to a vector of features. Cuay\'{a}huitl et al. \cite{Cuayahuitl05} used a hidden Markov model (HMM) for user simulation. The model generated both user and system actions. Schatzmann et al. \cite{Schatzmann07a} proposed a new agenda-based approach that did not necessarily need training data but could be trained in case such data was available \cite{Schatzmann07b}. Chandramohan et al. \cite{Chandramohan:11} proposed to model the user as a decision-making agent and to model user behaviour with reinforcement learning.

An important feature for user simulation, which encourages coherent behaviour throughout a dialogue, is the ability to take into account the dialogue history. For tractability reasons, previous models do not account for a long dialogue history. Another important consideration is that users who interact with SDS often utter several dialogue acts during a single dialogue turn. This feature is not often represented in user simulators, as it would quickly become inefficient to compute a model with an output space containing all possible sequences of dialogue acts. To deal with this, in the agenda-based approach, a stack-like structure is added to the model to provide a coherent set of dialogue acts that can be output at the same time. In the next section, we propose a model that takes into account the entire dialogue history and outputs a sequence of dialogue acts without relying on any external structure.

\section{The Sequence-to-sequence Model}
\label{seqtoseq}
\begin{figure}[!t]
\begin{center}
		\includegraphics[trim={3cm 7cm 2cm 2cm},scale = 0.3]{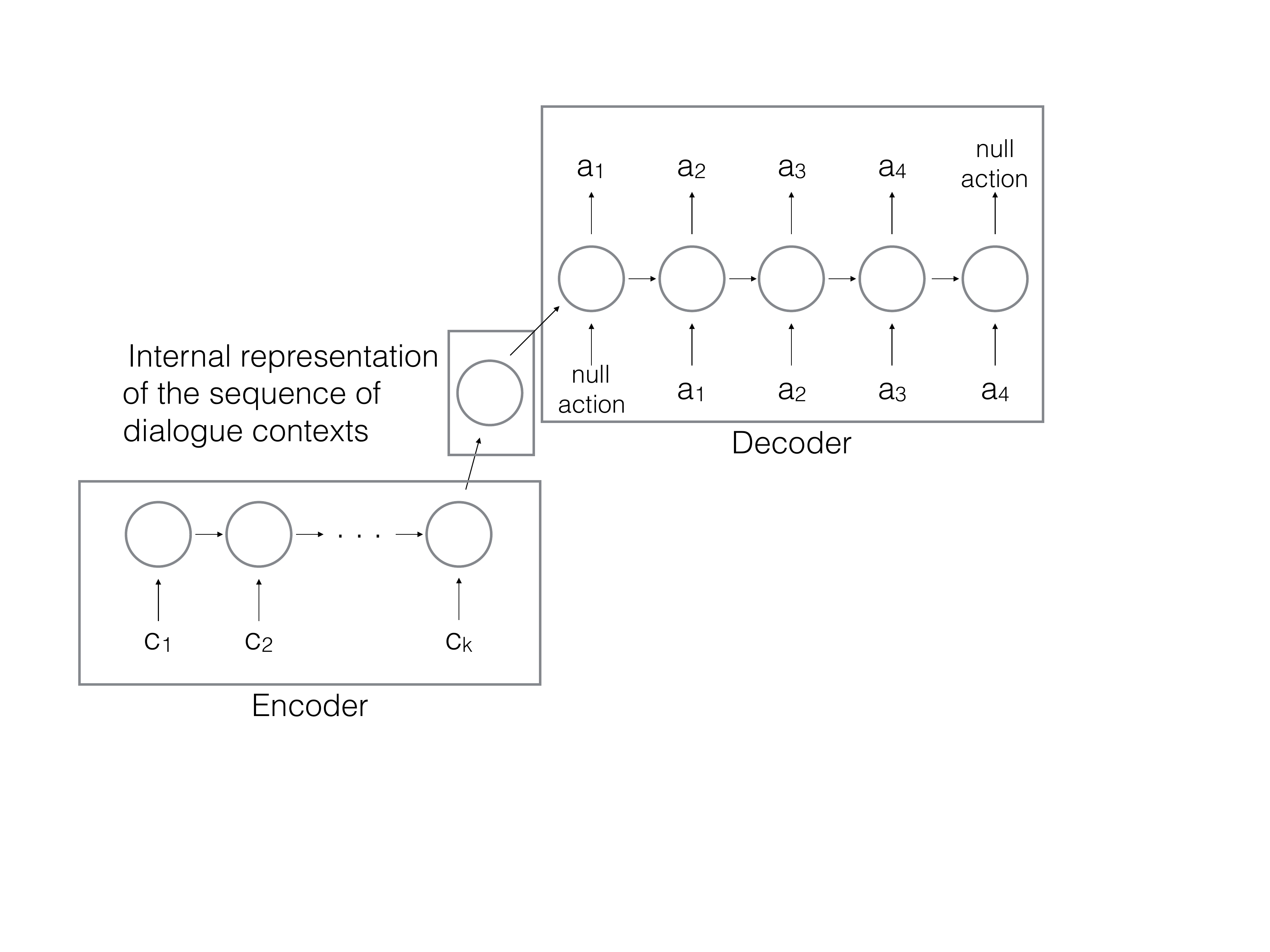}
	\caption{Sequence-to-sequence model for user simulation.}
	\label{fig:seqtoseq}
\end{center}
\vspace{-0.8cm}
\end{figure}

Figure \ref{fig:seqtoseq} represents the sequence-to-sequence user simulation model. The model takes as input a sequence of dialogue contexts $(c_1, c_2, ..., c_k)$ and outputs a sequence of actions $(a_1, a_2, ..., a_l)$.

Similarly to Schatzmann et al. \cite{Schatzmann07a, Schatzmann07b}, at the beginning of each dialogue, we uniformly draw a goal $G = (C, R)$ where $C$ is a set of \textit{constraints} and $R$ is a set of \textit{requests}. For a restaurant-seeking system, constraints are typically expressed over the type of food, the price range, and the area where the restaurant is located. Requests can include these slots as well as the restaurant's name, its address, its phone number, \textit{etc}.

A context $c_t$ at turn $t$ is defined by the following components:
\begin{itemize}
\item the most recent machine acts $a_{m,t}$,
\item the inconsistency between the most recent information provided by the machine and the user goal $\inc_t$,
\item the constraints status (informed or not) $\con_t$, and
\item the requests status (informed or not) $\req_t$.
\end{itemize}

The machine acts are encoded as a vector $a_{m,t}$ of size $n_{ma}$\footnote{where $n_{ma}$ is the number of possible machine acts.}. The vector $a_{m,t}$ has ones for the current machine acts and zeros everywhere else. The inconsistency is composed of two vectors whose size equals the number of possible constraints $n_c$. Both vectors are initialized at 0 and reset after each turn. After the machine makes a proposition to the user (\textit{e.g.}, a proposition of restaurant), all of the constraint slots which are in the user goal but which were not mentioned by the machine are set to 1 in the first vector. Every time the machine mentions a slot provided by the user (\textit{e.g.}, in a confirmation or a proposition), all of the constraint slots which have been misunderstood are set to 1 in the second vector. 
The inconsistency vector is thus a turn-level vector which models the system's understanding of the user goal. The constraints status vector is of size $n_c$ and keeps track of what the user has said to the machine. The constraints which are not in $C$ are set to 1 and those in $C$ are set to 0. Every time the user provides a constraint to the SDS, this constraint is set to 1. A constraint is reset to 0 every time it is set to 1 in the inconsistency vector or if the machine requests this slot. The requests status vector is of size $n_r$ where $n_r$ is the number of possible requests. This vector has ones for all slots which are not in the user goal and zeros for the slots in $R$. A request slot is set to 1 every time the SDS mentions it in a proposition. The requests status vector is reset after each new proposition from the system. Examples of updates are given in Table \ref{tab:ex_contexts}.
At time $t$, the sequence-to-sequence model takes as input the entire sequence of contexts that have been observed so far, which models dialogue history. This input is passed to an RNN which outputs a single vector $v_t$ corresponding to the model's internal representation of dialogue history. The encoder and the decoder have similar structures. They are both based on a Long Short-Term Memory (LSTM) \cite{Hochreiter:97}. In both cases, the LSTM is followed by a fully connected layer. The input of the encoder is a sequence of contexts $c_t$:
\begin{align}
c_t = a_{m, t} \odot \mathit{inconsist}_t \odot \mathit{const}_t \odot \mathit{req}_t \nonumber,
\end{align}
where $\odot$ is concatenation.
The LSTM are implemented following these equations:
\begin{align}
i_t &= \sigma(W_i c_t + U_i h_{t-1}) \nonumber \\
f_t &= \sigma(W_f c_t + U_f h_{t-1}) \nonumber \\
C_t &= i_t * \tanh(W_c c_t + U_c h_{t-1})  + f_t * C_{t-1} \nonumber \\
o_t &= \sigma(W_o x_t + U_o h_{t-1}) \nonumber \\
h_t &= o_t * \tanh(C_t),
\label{eq:lstm}
\end{align}
where $i_t$ is the input gate, $\sigma$ is the sigmoid function, $f_t$ is the forget gate, $o_t$ is the output gate, $C_t$ is the cell gate and $h_t$ is the hidden state. 
The last output of the LSTM is passed to one layer fully connected which outputs $v_t$. Then, $v_t$ is used to initialize the decoder LSTM at each time step \cite{Cho:14}. During training, the decoder is fed with ground truth, \textit{i.e.}, the sequences of user acts observed in the dataset given the history of contexts. During runtime, the only input to the decoder is the \textit{null} action. The decoder is implemented according to the same equations as the encoder. It is followed by \textit{softmax} activation in order to compute a distribution of probabilities over the actions. The first action ${a_{t,1}}$ is drawn according to the output distribution of the first step of the LSTM. Then it is fed as input to the second step. This process is repeated until a sequence of $l$ actions $(a_{t,1},...,a_{t,l})$ (including one or more \textit{null} actions at the end of the sequence) has been generated. We train this model with a categorical cross-entropy loss function.

Each sequence output by the simulator is a sequence of dialogue acts, \textit{e.g.}, \texttt{(inform, request)}. We map these dialogue acts to actions such as \texttt{inform(type of food = Chinese), request(price range)} by looking at the current user goal and uniformly drawing among the constraints left to inform and the requests left to ask. In the case of a confirmation asked by the system or if the system misunderstood a slot, we map the \texttt{inform} dialogue act to the slot in question. We show in the following section that it is also possible to train the model on original actions directly, \textit{e.g.}, \texttt{request-area}, which removes this post-processing step and models user behaviour at a finer level.

\section{Experiments}
\label{expes}
In this section, we compare the sequence-to-sequence simulator to an agenda-based simulator, a sequence-to-one model, and an n-gram model. We train these models on the training set of DSTC2.

We define a \textit{user compound act} $\tilde{a}^u_t$ as a sequence of dialogue acts $(a^u_{t,1},...,a^u_{t,l})$, where $l \geq 1$. All the models compared in this section output user compound acts. Similarly, we define machine compound acts as $\tilde{a}^m_t$.

\subsection{User Simulation Models}
The first baseline for comparison is a simple bi-gram model, which outputs a compound act $\tilde{a}^u_t$ given the last machine compound act $\tilde{a}^m_t$. We compute probabilities for the 54 possible user compound acts in the DSTC2 dataset.

In the agenda-based model, the user is modelled with a pair $(G, A)$, where $G$ is the goal and $A$ is the agenda. As explained in Section \ref{seqtoseq}, the goal is a pair $(C, R)$, where $C$ is a set of constraints and $R$ is a set of requests. The agenda $A$ is a stack-like structure which contains all of the \texttt{inform} and \texttt{request} acts needed by the user in order to perform her goal.\footnote{If the user is looking for an Indian restaurant downtown and wants to know the price range, the agenda will be: \texttt{inform(food = Indian)}, \texttt{inform(area = downtown)}, \texttt{request(price range)}.} At each dialogue turn $t$, the user simulator samples a single act $a^u_t$ based on the current dialogue context $d_t$\footnote{Since the dialogue contexts are not expressed in the same way for the sequence-to-sequence model and the agenda-based model, we use different notations.}. Then, based on the chosen act $a^u_t$, the user simulator samples the number $n$ of acts to pop from the stack. The compound act $\tilde{a}^u_t$ is then formed by $a^u_t$ and the acts that are popped from the stack. The dialogue context $d_t$ does not only include the latest dialogue acts spoken by the system, it also includes information on the dialogue history. For instance, if the SDS proposes a restaurant to the user and, in another dialogue turn, answers one of the user's requests regarding this restaurant, $d_t$ will include an indication over the goal status for this restaurant. The dialogue contexts combined with the agenda guarantee coherent user behaviour throughout the dialogue. This feature, as well as the fact that the model outputs one or several dialogue acts at each turn, makes this a good model for comparison with the sequence-to-sequence approach.

The third simulator is a sequence-to-one model. This model takes the same input as the sequence-to-sequence model but only outputs a probability distribution over a predefined set of compound acts. This set of size 54 contains all of the compound acts in DSTC2.

\subsection{F-score}
We compare the 4 models based on F-score. The F-score is the geometric mean of the precision and the recall, which are computed as follows:
\begin{align}
\text{precision} &= \frac{\text{number of correctly predicted dialogue acts}}{\text{number of predicted dialogue acts}} \nonumber \\
\text{recall} &= \frac{\text{number of correctly predicted dialogue acts}}{\text{number of dialogue acts in the corpus}} \nonumber \\
\text{F-score} &= 2 \times \frac{\text{precision}\times\text{recall}}{\text{precision} + \text{recall}}.\nonumber
\label{eq:f-score}
\end{align}

\begin{table*}[!t]
\begin{center}
\begin{tabu}to 1\linewidth{@{}X[1.5,c]X[0.8,c]X[c]X[1.2,c]X[1.5,c]@{}} \toprule
Dataset & Bigram & Agenda-based & Sequence-to-one & Sequence-to-sequence \\ \midrule
DSTC2 Validation & 0.20 & 0.24 & 0.37 & 0.34 \\
DSTC2 Test & 0.09 & 0.18 & 0.29 & 0.27 \\
DSTC3 Test & | & 0.13 & 0.19 & 0.18 \\
\bottomrule
\end{tabu}\vspace*{-2mm}
\end{center}
\caption{Average F-score on 50 runs.\vspace*{-2mm}}
\label{tab:res}
\end{table*}

\subsection{The DSTC2 dataset}
DSTC2 is a publicly available dataset composed of a training set of 1612 dialogues, a validation set of 506 dialogues and a test set of 1117 dialogues. The training and validation sets were collected with two handcrafted policy managers whereas a statistical policy manager was used for the test set. The dialogues were collected with real users who had been given a goal consisting of a set of constraints and a set of requests. Each user interacted with the system in order to find a restaurant matching all of the constraints and then to collect the information in the requests. The user dialogue acts tagged in this dataset are as follows: \texttt{deny}, \texttt{null} (empty act), \texttt{request more}, \texttt{confirm}, \texttt{acknowledge}, \texttt{affirm}, \texttt{request}, \texttt{inform}, \texttt{thank}, \texttt{repeat}, \texttt{request alternative} (ask for another option), \texttt{negate}, \texttt{goodbye}, \texttt{hello} and \texttt{restart} (ask the system to restart the dialogue).

This dataset offers an interesting setting since we can use both the validation and test sets in order to evaluate the user simulators. In general, a user simulator is designed for a given policy manager: data is collected with this manager then the user simulator model is trained on this data and evaluated with the same policy manager. With this dataset, we have the possibility to follow this methodology (on the validation set) but we are also able to evaluate on a set of dialogues on the same domain but collected with a different policy manager (the test set). Therefore, we can evaluate the extent to which each model captures the behaviour of real users in unseen settings for the same task.
\subsection{Results}
Table \ref{tab:res} presents results on the validation and test sets of DSTC2. The first observation is that, as expected, the bi-gram model performs relatively poorly. On both the validation and test sets, the RNN-based models significantly outperform the agenda-based model in terms of F-score. The sequence-to-one model performs slightly better than the sequence-to-sequence model because it is a simpler problem to learn a distribution over a given set of sequences than to output each sequence step by step. However, the sequence-to-sequence model performs very closely to the sequence-to-one simulator, demonstrating that this model can achieve good performance. In addition, a considerable advantage of this model concerns scalability. In particular, the number of possible compound acts might grow considerably if the sets of constraint and request slots were of larger size and/or if the number of dialogue acts was larger. The output space would rapidly become too large for training the sequence-to-one model on a small dataset and it would likely be more efficient to use the sequence-to-sequence model. A further advantage is that the sequence-to-sequence model can be used on the original act space.

We illustrated this property with a second experiment, in which we modify the sequence-to-sequence model to train it on the original action space. The dialogue acts generated by the simulator are uniformly mapped to the user goal as discussed in Section \ref{seqtoseq}. In this experiment, we circumvent this random mapping by increasing the number of possible acts. Instead of having one \texttt{inform} dialogue act, we define three separate acts: \texttt{inform\_food}, \texttt{inform\_pricerange}, \texttt{inform\_area}. The advantage of this format is that a mapper is no longer needed and users can be modelled at finer granularity. Indeed, as shown in Table \ref{tab:res_original}, it is possible to learn the order in which constraint and request slots are provided to the system by users. For instance, in the case that the user goal includes food, area and price range, the encoder-decoder model learns, in proportions commensurate with those found in the corpus, that the food slot is most often preferred as the first slot (72\% in the corpus, 48\% for the simulator), then the price range (16\% vs. 31\%), and then the area (12\% vs. 21\%).

\begin{table}
\begin{center}
\begin{tabu}to\columnwidth{@{}X[2,c]X[c]X[c]X[3,c]@{}} \toprule
Slot & in goal & corpus & sequence-to-sequence \\ \midrule
area & yes & 6 & 23.1 \\
price range & no & 0 & 10.4 \\
food & yes & 140 & 221.0 \\\midrule
area & yes & 15 & 101.8 \\
price range & yes & 19 & 153.1 \\
food & yes & 86 & 238.2 \\
\bottomrule
\end{tabu}
\end{center}
\caption{For two different user goals, we compute the count of when a slot has been the first to be provided to the system in the corpus and by the sequence-to-sequence model (averaged over 10 runs). Note that when the user informs the system of a slot which is not in the goal, the value for this slot is \textit{do not care}.}
\label{tab:res_original}
\end{table} 

The last experiment involves evaluating the simulators on the DSTC3 test set \cite{Henderson:14c}. DSTC3 is a dataset of 2264 dialogues with a system that can search for restaurants, pubs and coffee shops. Compared to DSTC2, in this dataset, the number of possible constraints is increased with the following slots: children allowed, has internet, has tv, near (\textit{e.g.}, nearby Queens college) and type (restaurant, pub or coffee shop). 
The user and system dialogue acts can easily be mapped to those in DSTC2. We use this dataset in order to evaluate the user simulators on a new, larger domain. We train the models on DSTC2 as before, and evaluate them on the DSTC3 test set based on F-score. The results are presented in Table \ref{tab:res}. These show that the sequence-to-one and sequence-to-sequence models significantly outperform the agenda-based model. Compared to DSTC2, there is a degradation in F-score which can be explained by the fact that this new domain has a larger set of compound acts (we found 40 compound acts which never occurred in DSTC2). The degradation concerns mostly the recall. Notably, the F-score for these models is similar to the F-score of the agenda-based model on the test set of DSTC2. 


\section{Conclusions}
\label{conclu}
We proposed a new sequence-to-sequence model for user simulation in spoken dialogue systems. Compared to previous models, this simulator takes into account the entire dialogue history, it does not rely on any external data structure to ensure coherent user behaviour, and it does not require mapping to a summarized action space, which makes it able to model user behaviour with finer granularity. We showed that this model outperforms a state-of-the-art simulator based on the F-score measure. We also showed that it can be efficiently transferred to a new information-seeking domain. In future work, we will use the model to train a statistical spoken dialogue system and further explore the potential of this architecture.

  \newpage
  \eightpt
  \bibliographystyle{IEEEtran}
  \bibliography{biblio}

\end{document}